# Shapes of Cognition for Computational Cognitive Modeling

**Marjorie McShane** MARGEMC34@GMAIL.COM
**Sergei Nirenburg** ZAVEDOMO@GMAIL.COM
**Sanjay Oruganti** SANJAYOVS.RPI@OUTLOOK.COM
**Jesse English** DRJESSEENGLISH@GMAIL.COM
Language-Endowed Intelligent Agents Lab, Rensselaer Polytechnic Institute, Troy, NY 12180, USA

Abstract

*Abstract. Shapes of cognition* is a new conceptual paradigm for the computational cognitive modeling of Language-Endowed Intelligent Agents (LEIAs). *Shapes* are remembered constellations of sensory, linguistic, conceptual, episodic, and procedural knowledge that allow agents to cut through the complexity of real life the same way as people do: by expecting things to be typical, recognizing patterns, acting by habit, reasoning by analogy, satisficing, and generally minimizing cognitive load to the degree situations permit. *Atypical* outcomes are treated using *shapes*-based recovery methods, such as learning on the fly, asking a human partner for help, or seeking an actionable, even if imperfect, situational understanding. Although *shapes* is an umbrella term, it is not vague: *shapes*-based modeling involves particular objectives, hypotheses, modeling strategies, knowledge bases, and actual models of wide-ranging phenomena, all implemented within a particular cognitive architecture. Such specificity is needed both to vet the our hypotheses and to achieve our practical aims of building useful agent systems that are explainable, extensible, and worthy of our trust, even in critical domains. However, although the LEIA example of *shapes*-based modeling is specific, the principles can be applied more broadly, giving new life to knowledge-based and hybrid AI.

## 1. Introduction

*Shapes of cognition* is a new conceptual paradigm for the computational cognitive modeling of Language-Endowed Intelligent Agents (LEIAs). *Shapes* are remembered constellations of sensory, linguistic, conceptual, episodic, and procedural knowledge that allow agents to cut through the complexity of real life the same way as people do: by expecting things to be typical, recognizing patterns, acting by habit, reasoning by analogy, satisficing, and generally minimizing cognitive load to the degree situations permit. *Atypical* outcomes are treated using *shapes*-based recovery methods, such as learning on the fly, asking a human partner for help, or seeking an actionable, even if imperfect, situational understanding.[1] A *shapes*-based architecture recognizes and reuses complex instances, not only general rules and first-principles reasoning, thereby reducing computational burden and increasing explanatory power (Gentner, 1983; Newell & Simon, 1972). A *shapes* orientation also helps to prioritize tasks in cognitive modeling and offers metrics for the evaluation of corresponding agent systems.

Although *shapes* is an umbrella term, it is not vague: *shapes*-based modeling involves particular objectives, hypotheses, modeling strategies, knowledge bases, and actual models of wide-ranging phenomena, all implemented within a particular cognitive architecture. Such specificity is needed both to vet the our hypotheses and to achieve our practical aims of building useful agent systems that are explainable, extensible, and worthy of our trust, even in critical domains. But although the LEIA example of *shapes*-based modeling is specific, the principles can be applied more broadly, giving new life to knowledge-based and hybrid AI.

---

[1] Shape-based modeling does not follow the practice of picking the low-hanging fruit in one task and then moving onto something else when the preferred methods (typically supervised machine learning) hit a ceiling of results. This practice was widely pursued in pre-LLM natural language processing, bolstered by fieldwide competitions.

As an example of how people orient around shapes in real life, consider the domain of clinical medicine. Whereas certain constellations of property values (symptoms, observations, diagnostic test results, etc.) are sufficient to definitively diagnose a disease, others can only support a particular hypothesis, whereas still others leave the space of options very broad. In other words, there is graded typicality that directly informs decision-making. Knowledge of the associated constellations of property values is not only useful to clinicians, it is required as formal justification for prescribing medications, ordering tests, and carrying out procedures. So, anticipating and orienting around what is typical is not lazy—it is how people manage in a remarkably complex world.

**Background:** *Shapes*-based modeling evolved in our work on developing LEIAs, which are neurosymbolic, multimodal, cognitive-robotic systems implemented in the HARMONIC architecture, shown in Fig. 1 (Oruganti et al. 2024a, 2024b). The cognitive (strategic) layer primarily relies on knowledge-based methods to support reliability and transparency, whereas the robotic (tactical) layer primarily relies on machine learning, which is effective and sufficient since the associated capabilities (e.g., the "how" of moving a robotic arm) need no explanation. The components of the cognitive and robotic layers function both independently and interactively.

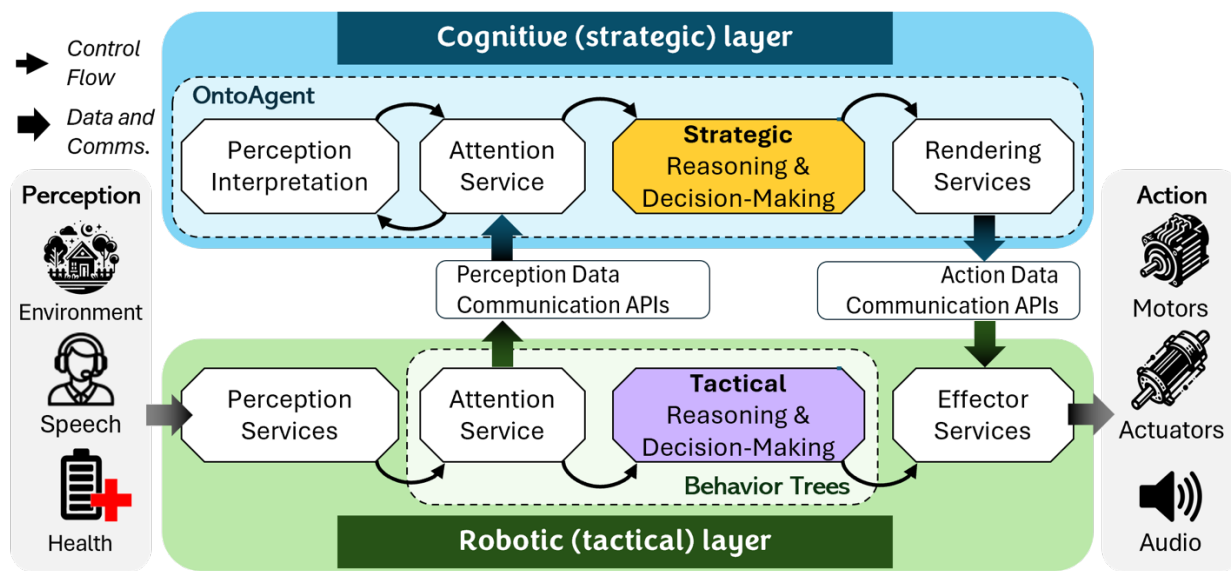

**Fig. 1.** The HARMONIC cognitive-robotic architecture.

LEIAs represent a novel approach to *Agentic AI*, which we call *OntoAgentic AI*. Whereas typical agentic AI systems use LLMs both as the orchestrator and for support functions, *OntoAgentic AI* uses a LEIA as the orchestrator and leverages both LEIA agents and LLM-based systems for support functions. OntoAgentic AI, therefore, offers reliable, explainable control of overall system operation as well as the cognitive operation of each individual LEIA. LEIA cognition orients around meaning, which is defined in terms of an unambiguous, language-independent ontology, following the theory of Ontological Semantics (Nirenburg & Raskin, 2004). When LEIAs interpret stimuli, reason, or plan, they do it in terms of ontologically-grounded meaning representations like the following, which expresses the idea that *Tony was watching a tiger*.

```
VOLUNTARY-VISUAL-EVENT-1
    AGENT              HUMAN-1
    THEME              TIGER-1
    TIME               < find-anchor-time
    ASPECT             progressive
    episodic-mem       VOLUNTARY-VISUAL-EVENT-#9
HUMAN-1
    HAS-NAME           Tony
    episodic-mem       HUMAN-#17
TIGER-1
    DISCOURSE-STATUS   new
    episodic-mem       TIGER-#1
```

Meaning representations are our first example of a *shape*: they are frame-based structures written in an unambiguous knowledge representation language that is well suited to agent reasoning. Ontological concepts are written in small caps to distinguish them from words of English. The plain numerical indices distinguish instances of concepts in the meaning representation. The numerical indices preceded by # indicate grounding to episodic memory; this is true *reference* resolution, in contrast to textual *co*reference resolution. The value of TIME must be computed dynamically, by first establishing the speech time ("anchor-time") and then understanding that the event occurred before (<) that.

Although it is difficult and expensive to enable agents to interpret their experiences, reason, and learn in terms of an ontological model, there are three main benefits to doing this: (1) agents can use all of the knowledge stored about ontological concepts in their reasoning; (2) all of their knowledge and traces of their reasoning are human-inspectable, which will lead to trust in agent systems; and (3) the vast majority of agent knowledge and reasoning is language-independent, which allows for LEIAs to be ramped up in any natural language with only a change in the language processors at the flanks of the cognitive architecture.

**Back to *shapes*:** We have long been thinking in terms of *shapes* when modeling LEIAs without promoting this idea to an overarching principle. There are four reasons for doing this now. (1) LEIAs' capabilities are rapidly expanding, and many of them—language processing, learning, decision-making under uncertainty, etc. —are not only individually of unbounded complexity, they are also interdependent. It takes effort and innovation to keep the functioning of such a system transparent, and *shapes* is one aspect of that innovation. (2) We have recently been transitioning from pure research to deploying cognitive-robotic systems. This necessitates (a) precisely characterizing what our agents can and cannot do at any given time, (b) planning development work toward specific, achievable goals, and (c) creating metrics for performance and evaluation. *Shapes* serve all of these objectives. (3) The boom of language models has led to more sci-fi hype and doomsday prognoses than concrete, alternative approaches to AI. This is a mistake: it is too early to be shutting down the conversation about methods. (4) The only way cognitively grounded, symbolic AI can survive in the current climate is if decision makers can easily understand what it is, how it works, and what benefits it offers. *Shapes* help to translate the science beyond the inner circle of likeminded developers.

*Shapes* is more inclusive than related phenomena in the literature, such as prototypes (Rosch, 1973), memory organization packets (Schank, 1982), frames (Fillmore & Baker, 2009), constructions (Hoffmann & Trousdale, 2013), chunking (Thalmann, Souza, & Oberauer, 2019), and templates (Sung et al., 2021). Shapes-based modeling underscores common objectives and features across the different cognitive models that are interleaved in an end-to-end agent system. To use an architectural metaphor, the cognitive models underlying LEIAs are like rooms in a well-designed, finely crafted home: despite differences in form and function, they show marked aspects of continuity in keeping with their shared blueprint.

Although shapes might seem like an obvious knowledge engineering anchor for readers schooled in linguistics and/or early AI, for those fully ensconced in statistical AI, the only shapes that might come to mind are nets and black boxes, which are *anti*-shapes. If nets and black boxes were enough, then the work of symbolic cognitive modeling would be strictly academic—not uninteresting, but exclusively to satisfy human curiosity. However, nets and black boxes are not enough. Although empirical AI offers many useful capabilities, it is not reliable or learning-enabled in the specific ways that many applications require. Reports about this are too numerous to list (e.g., Connell & Lynott, 2024; Cuskley, Woods and Flaherty, 2024; Kapoor et al., 2024; Xu, Jain & Kankanhalli, 2024).

This paper gives an example-based taste of how the cognitive (strategic) side of LEIA modeling is all about shapes, and how shapes foster the science, the technology, and the dissemination of this program of R&D. Specifically, we briefly describe three shapes of knowledge—involving lexicon, ontology, and episodic memory, and three shapes of reasoning—involving language understanding, language generation, and learning ontological scripts. Six more examples of shapes are noted in the Discussion section. Two recent books that are widely referenced below will be cited using abbreviations: *Linguistics for the Age of AI* (McShane & Nirenburg, 2021) is LingAI, and *Agents in the Long Game of AI* (McShane, Nirenburg, & English, 2024) is LongGame.

## 2. Shapes of Lexicon

Human-oriented lexicons and syntactic theories of language classify word senses according to constructions like transitive, intransitive, and ditransitive. Our approach to lexical storage, called *construction semantics*, takes constructions to a new level by precisely specifying the aligned syntactic and semantic expectations of word senses (see LongGame, sections 3.3 and 4.2.2).[2]

As an example, Table 1 shows three lexical senses that have the same semantic shape, recorded in the sem-struc zone (ADMIRE with an AGENT and a THEME), but different syntactic shapes, recorded in the syn-struc zone. Variables indicate cross-references, and ^ indicates "the meaning of" the variable. The class names for the syntactic and semantic shapes are values of the syn-class and sem-shape fields, respectively.

**Table 1.** Three constructions with the semantic shape (in boldface) but different syntactic shapes.

| admire-v1 | look-v24 | put-v29 |
|---|---|---|
| ex: John admires his uncle. | ex: John looks up to his uncle. | ex: John puts his uncle on a pedestal. |
| syn-class: v-trans | syn-class: v-part-pp | syn-class: v-do-pp |
| sem-shape: EVENT(AGENT,THEME) | sem-shape: EVENT(AGENT,THEME) | sem-shape: EVENT(AGENT,THEME) |
| syn-struc<br>   subject    $var1<br>   v           $var0<br>   directobject $var2<br>sem-struc<br>   ADMIRE<br>      AGENT   ^$var1<br>      THEME   ^$var2 | syn-struc<br>   subject    $var1<br>   v           $var0<br>   part       up<br>   pp<br>      prep    to<br>      obj     $var2<br>sem-struc<br>   ADMIRE<br>      AGENT ^$var1<br>      THEME ^$var2 | syn-struc<br>   subject    $var1<br>   v           $var0<br>   directobject $var2<br>   pp<br>      prep    on<br>      obj<br>         det   a<br>         n     pedestal<br>sem-struc<br>   ADMIRE<br>      AGENT   ^$var1<br>      THEME   ^$var2 |

---

[2] Construction semantics differs from construction grammar (Hoffmann & Trousdale, 2013) in that it is computational rather than theoretical, it treats semantics as centrally as syntax, and it grounds meaning in a formal ontology.

Apart from faithfully capturing what people know about the meaning and usage of words and phrases, lexical shapes offer several practical benefits:

- Shapes help agents to learn new word senses on the fly: the syntactic parse identifies the syn-struc shape, which constrains and prioritizes the options for the sem-struc shape (*LongGame*, ch. 7).
- Shapes are templates that speed up lexical acquisition by people, which is needed for difficult cases (*LongGame*, ch. 9).
- Development, testing, and evaluation of the natural language understanding and generation systems (cf. Sections 5 and 6) can be done at the level of shapes rather than exhaustively. For example, if a given syntactic transformation, such as passivization, works on a verb sense of the syntactic-class "v-part-pp" (cf. look-v24 in Table 1), then it should work on other verb senses in that class, apart from exceptions in language or parsing errors.
- Classifying lexical senses according to shapes allows for more insightful assessments of progress than simple counting, since a sense can range from simple (a noun mapped to a corresponding concept) to mid-complexity (a verbal idiom like *Subj$_1$ take it upon Refl-pro$_1$ to* VP) to requiring a specialized procedural semantic routine (as to compute how *very* modifies the value of scalar attributes).

In the general case, only the basic shape of argument-taking words is explicit in the lexicon, such as the active voice of verbs. When a word is used in another way—e.g, in the passive voice, as an infinitival complement of another verb, or in a question— dynamic transformations are needed. We describe those with respect to language understanding and generation in Sections 5 and 6, respectively.

## 3. Shapes of Ontology

LEIAs' knowledge of the world, in terms of types, not instances, is recorded as a frame-based graph of concepts organized as an inheritance hierarchy.[3] A small excerpt from the concept SURGERY is shown below.

| SURGERY | | | |
|---|---|---|---|
| | IS-A | **value** | **MEDICAL-PROCEDURE** |
| | AGENT | **default** | **SURGEON** |
| | | sem | PHYSICIAN |
| | | relaxable-to | HUMAN, ROBOT |
| | LOCATION | **default** | **OPERATING-ROOM** |
| | | sem | MEDICAL-BUILDING |
| | | relaxable-to | PLACE |

Frames themselves are *shapes,* but *shapes* play out in other interesting ways in the ontology as well; these involve facets, inheritance, and scripts.

Facets reflect different strengths of constraints, shown above using different font styles. Whereas, by default, a surgeon performs surgery in an operating room (reflecting the strongest, most expected constraints), any physician can perform some kinds of surgery in any medical building, and, in a pinch, a regular person might perform some lifesaving invasive procedure anywhere (e.g., a soldier on a battlefield). So, whereas property-value information in concept descriptions captures expectations about the world, facets make those expectations ever more precise.

---

[3] Ontologies intended for AI are not necessarily organized as frames or inheritance hierarchies. For example, although the heavily funded Cyc ontology originally used a frame-like architecture, the knowledge representation strategy shifted to a "sea of logical assertions," with each assertion being equally about each of the terms used in it (Mahesh et al., 1996, p. 21). For further discussion, including why we don't use Cyc, see *LongGame* section 3.1.

Another way that the shapes are reflected in the ontology is that knowledge is organized as an inheritance hierarchy. There are multiple reasons for having an inheritance-based ontology: it allows for agents to reason about subclasses and superclasses; it fosters both manual knowledge acquisition and agent learning, since only locally distinct property values need be specified; it facilitates agent reasoning about the salient distinctions between proximate concepts in the ontological space, since these are the ones that are locally specified; and it facilitates broadscale knowledge acquisition by making clear which properties are most salient within a given subtree and, therefore, need to be locally specified for all concepts.

The final *shapes*-oriented aspect of the ontology that we will mention here involves scripts (also called complex events or procedural knowledge), which are sequences of events with their participants and props (Schank & Abelson, 1977). They record how complex events typically play out, along with options, coreferenced variables, loops, and the rest. They reflect the skeleton of a complex action that then fills out in a particular way when realized as a plan. Scripts allow agents both to make plans and to recognize when others are carrying out a plan. They are recorded in frames like those above but enhanced by additional expressive means, for example to indicate coreferences among slot fillers.

In short, the ontology reflects how the world typically works in terms of knowledge structures whose shapes are readily interpretable by LEIAs, human developers, and others who want to understand the knowledge substrate that guides LEIA operation. As concerns evaluation, shapes offer the same kinds of benefits for the ontology as for the lexicon in that they allow for enhancements to be characterized at a fine grain-size, and they allow glassbox system evaluations to identify needs for knowledge enhancement.

## 4. Shapes of Episodic Memory

Episodic knowledge is recorded as instances of ontological concepts using the same frame-based structures as the ontology. So, the shape of a memory about a particular object or event mirrors the shape of the concept that it instantiates. This means that agents can reason about elements of episodic knowledge by consulting the concept in the ontology, along with its full property-based description. Below is an excerpt from a remembered instance of SURGERY: a particular SURGEON (indexed as #14 in the agent's memory) operated on a particular PATIENT (#89) in a particular HOSPITAL (#3) on December 12, 2024.

SURGERY-#10
    AGENT         SURGEON-#14
    BENEFICIARY  PATIENT-#89
    THEME         APPENDIX.PART-OF.PATIENT-#89
    LOCATION     HOSPITAL-#3
    DATE          2024-12-12

Each of the property fillers (SURGEON-#14, PATIENT-#89, HOSPITAL-#3) also has its own property-rich description in episodic memory.

Here we focus on three uses of shapes in episodic memory: reasoning by analogy, memory consolidation, and planning that incorporates others' personal preferences.

Reasoning by analogy enables agents to circumvent reasoning from first principles by repeating something that worked in the past (e.g., Gentner & Smith, 2013). For example, if an agent needs to create a plan from an option-filled ontological script, the least-effort strategy is to copy the last plan that worked (or some other previous plan that worked well or frequently) unless it is infeasible in the given circumstances. Similarly, if the agent receives an ambiguous input, but its past analyses of an identical or similar input resulted in a particular interpretation, then the least-effort action is to interpret the new instance in the same way. For example, if an agent's human collaborator has formerly said "I need a cup of coffee" as a signal that he's about to take a break (not as a request that the agent get him one), then the agent can select that interpretation without extensive reasoning or the need to initiate a clarification dialog.

Of course, matching in service of analogical reasoning can be tricky: *how* similar and *in what ways* do past and current meaning representations (situations, inputs, decisions, etc.) need to be to warrant reasoning by analogy? This could spiral into endless complexity but a *shapes*-based orientation helps. The fact is, LEIAs don't need to able to reason by analogy at a human level in order to use analogy as a useful tool. We can prepare them to reason by analogy in specific ways in specific kinds of situations. If they recognize a shape (constellation of feature values) that enables them to reliably reason by analogy, then they do. If not, they use some other reasoning strategy. Of course, this requires that we, as developers, specify exactly how we want reasoning by analogy to work, without offloading it to the seemingly magic but unreliable operation of language models. Explicit cognitive modeling of this sort contributes both to cognitive science and to the development of reliable agent systems.

Another manifestation of shapes in episodic memory involves consolidating memories of regularly repeating events. For example, if the agent observes multiple instances in which its human partner, Lou, puts his hammer, screwdriver, or wrench back in his toolbox right after he uses it, then this is best consolidated into the fact that Lou always puts back his tools after he uses them. As with reasoning by analogy, we must give LEIAs specific reasoning procedures to identify common *shapes* (repeating constellations of property values) that suggest habits. These procedures are attached to the concept IDENTIFY-HABIT. There is a similar concept, called CREATE-HABIT, whose functional attachment guides the agent in morphing its own deliberative action (involving System 2 reasoning) into a habitual one (involving System 1 reasoning) (see, e.g., Sun, Slusarz & Terry, 2005, for more on the two-system view). As with reasoning by analogy, agents can usefully recognize and develop habits based on common *shapes* long before achieving human-level competency.

A final use of shapes in episodic memory involves the agent's memory of how different humans prefer to carry out tasks. For example, two different people might teach the agent to carry out a complex action in two different ways. The agent will record both of these as options in its ontological script for the action, but it will record the different path preferences as episodic knowledge about each collaborator. Effectively, this episodic knowledge is like different stencils over the full script, providing strong guidance for how the agent should create its plan when working with each collaborator.

We have just seen some examples of how *shapes* play out in an agents knowledge bases. Now we turn to their use in agent reasoning.

## 5. Shapes of OntoSyntax and Basic Semantic Analysis

Language understanding by LEIAs is a six-stage process that involves multiple stages of syntactic analysis followed by multiple stages of semantic and pragmatic analysis, as detailed in LingAI and LongGame. The two consecutive stages we will discuss here are OntoSyntax and Basic Semantics, which together select which combinations of lexical senses can be used to analyze an input sentence. For example, given the input *Tony was watching a tiger,* OntoSyntax determines that the first verbal sense of watch, watch-v1 (Table 2) is syntactically compatible with the input (both are transitive), and Basic Semantics determines that this sense is semantically compatible with the input (*Tony* is an ANIMAL and *a tiger* is a PHYSICAL-OBJECT).[4]

---

[4] The semantic constraints are written in gray because they are not written explicitly in the lexicon; they are drawn from the ontology during language analysis and generation.

**Table 2.** Watch-v1 can be used to analyze *Tony was watching a tiger,* resulting in the meaning representation to the right.

| watch-v1 | | | VOLUNTARY-VISUAL-EVENT-**#9** | |
|---|---|---|---|---|
| syn-struc | | | **AGENT** | **HUMAN-#17** |
|   subject | $var1 | | **THEME** | **TIGER-#1** |
|   v | $var0 | | TIME | < find-anchor-time |
|   directobject | $var2 | | ASPECT | progressive |
| sem-struc | | | lex-map | watch-v1 |
|   **VOLUNTARY-VISUAL-EVENT** | | | | |
|     **AGENT** | **^$var1** | (sem ANIMAL) | | |
|     **THEME** | **^$var2** | (sem PHYSICAL-OBJECT) | | |

However, not all examples are so simple because word senses can be used in non-basic ways as well. For example, the sentence *Mary needed to feed Spot before going out to dinner* includes three verbs whose basic forms are recorded in the lexicon as shown in table 3.

**Table 3.** Three lexical senses needed to understand *Mary needed to feed Spot before going out to dinner.*

| need-v2 | | | feed-v1 | | | go-v54 | |
|---|---|---|---|---|---|---|---|
|  def | need plus an xcomp | |  def | To give food to; transitive | |  def | phrasal: X go out to dinner |
|  ex | I needed to do my homework | |  ex | She fed the dog | |  ex | We're going out to dinner. |
| syn-struc | | | syn-struc | | | syn-struc | |
|   subject | $var1 | |   subject | $var1 | |   subject. | $var1 |
|   verb | $var0 | |   verb | $var0 | |   verb | $var0 |
|   xcomp | $var2 | |   directobject | $var2 | |   prep-part | out |
| sem-struc | | | sem-struc | | |   pp | |
|   MODALITY | | |   FEED | | |     prep | to |
|     TYPE | OBLIGATIVE | |     AGENT | ^$var1 | |     obj | dinner |
|     VALUE | 1 | |     BENEFICIARY | ^$var2 | | sem-struc | |
|     SCOPE | ^$var2 | | | | |   EAT-AT-RESTAURANT | |
|     ATTRIBUTED-TO | ^$var1 | | | | |     AGENT | ^$var1 |
|   ^$var2 | | | | | | | |
|     AGENT | ^$var1 | | | | | | |

In our sentence, the first verb, *need*, is used in its basic form (all of its expected syntactic dependencies are accounted for), but the others, *feed* and *go*, are not. *Feed* needs to be converted into an infinitive clause and its missing subject needs to be understood as coreferential with that of *need;* and *go* needs to be converted into a present participle and its missing subject needs to be understood as coreferential with that of *need and feed*.

    In the tradition of generative grammar, dynamic modifications to recorded lexical knowledge are called transformations (Chomsky, 1957). But, whereas generative grammar considers only the syntactic aspect of transformations, LEIAs need to carry along the semantic interpretations as well.[5]

    However, having transformations in an agent's toolbox does not mean that nothing apart from the most atomic knowledge should be stored in the lexicon. Rather, there are both psychological and computational reasons for storing some complex, frequently-encountered entities explicitly, even if they are not idiomatic

---

[5] Of course, it is a hypothesis that lexical knowledge is divided into static and transformational components. Some threads of a competing theory of human language processing, construction grammar, reject the notion of transformations (Hoffmann & Trousdale, 2013). However, the latter do not specify the actual content of the lexicon precisely enough to offer an alternative that would be suitable for computational modeling. We find transformations both psychologically plausible and computationally useful, so they are used for both language understanding and language generation by LEIAs.

(i.e., semantically non-compositional). Psychologically speaking, storing frequent complex constructions minimizes effort, speeds processing, and seems to be what people do (e.g., Bybee, 2013). Computationally speaking, although one can implement individual transformations and apply them in sequence, this can get complicated both in terms of ordering (*is* there a fixed order that works for every possible combination of transformations?) and in terms of anticipating all of the non-basic structures that might serve as input to the *nth* transformation in the sequence (for details, see *LongGame*, Section 4.2.2).

Therefore, the LEIA's lexicon includes many complex constructions that offer stored interpretations rather than relying on runtime composition. An example is *what-interrogpro7* which combines a wh-question about the object of the main verb with passivization of that verb. This structure is presented using a commented shorthand for reasons of space.

```
what-interrogpro7
    def     a "what" question asking about a passivized DO
    ex      What was Mary given (by the workers)?
    syn-struc
        What_DirectObj  was  [NP_IndirectObj]1  [Verb_Bare-Past-Participle]2  (by  [NP_ObjOfPrep]3)?
    sem-struc
        REQUEST-INFO-WHAT-THEME
            THEME           ^$var2.THEME           ; "what" is the THEME of the main verb
        $var2
            (AGENT          ^$var3 )               ; optional in the input
            BENEFICIARY     ^$var1
            TIME            < find-anchor-time
```

Note that this semantic description—the exact shape of the sem-struc—will be reused in other lexical senses that are paraphrases of this question, such as *What did the workers give Mary?, What was it that the workers gave Mary?* and *What was it that was given to Mary by the workers.*

Let us recap how *shapes* helps to organize our thinking about automatic semantic analysis and its development, evaluation, and dissemination. Semantic analysis is informed by the syntactic parse, which is in the shape of a tree. That shape must either directly align with the syntactic shapes (syn-strucs) of argument-taking words in the lexicon or dynamic transformations must account for the discrepancies. If a particular input is not properly understood by the semantic analyzer, possible fail points are the lack of a lexical sense of the needed shape or the lack of a transformation to convert an existing lexical sense into the shape needed by the input. Knowledge engineers and software engineers jointly determine—through a combination of introspection and testing—how robust dynamic transformations can be, and when it is better to record complex constructions as explicit shapes in the lexicon.

## 6. Shapes of Meaning for Language Generation

Just as language understanding is a multi-step process, so, too, is language generation. First the agent recognizes the need to say something; then it formulates the meaning it wants to express using an ontologically-grounded meaning representation; and finally it decides how to express it using the word senses stored in its lexicon (LongGame, Section 4.3).[6] Here we focus on *shapes of meaning,* which help with that last step: translating meaning representations into sentences. This poses the same transformation-oriented challenges as we just saw with language understanding.

*Shapes of meaning* are variable-inclusive templates of ontological concepts that we hypothesize scaffold human thought just as linguistic constructions scaffold human languages. For example, just as the sentence

---

[6] Actually, the agent can also generate language reflexively, as to shout "Fire!" when it perceives fire, but that goes beyond the scope of this paper.

"Who do you think wanted to be selected?" contains nested constructions that are part of a person's knowledge of English (a wh-question scoping over a verb that takes an infinitival complement), the *meaning* of that sentence reflects nested frames of ontological concepts that are independent of any natural language. We computationally model the process of language generation by formalizing the links between shapes of meaning and their corresponding linguistic constructions.

Since we just discussed language understanding, the obvious question is, *Why aren't shapes of meaning used for language understanding?* The reason is not theoretical, it is practical. For language analysis, high-quality data-driven syntactic parsers exist and it would be a poor use of resources to reinvent syntactic parsing due to a rigid commitment to human-inspired cognitive modeling. After all, AI is a field that combines science with technology, and it needs to serve practical ends. But since no empirically grounded "reverse parser" exists for generation, we have to handle more low-level computational details explicitly, as part of cognitive modeling.

The key insight of *shapes of meaning* is that the overall shape of the meaning to be expressed (i.e., the full proposition) affects how the individual components need to be expressed. Consider the examples in Table 4, all of which include the information that a bicycle is blue. Depending on what else is included in the meaning representation, the bicycle's blueness is expressed variously in English.

**Table 4.** Different meaning representations and renderings involving a bicycle's blue color, with time and aspect removed for clarity of presentation.

| BICYCLE-#2<br>    COLOR   blue | COST-#1<br>    DOMAIN   BICYCLE-#2<br>    RANGE   .8<br>BICYCLE-#2<br>    COLOR   blue | AMUSE-#1<br>    CAUSED-BY   COLOR-#1<br>    EXPERIENCER   HUMAN-#1<br>COLOR-1<br>    DOMAIN   BICYCLE-#3<br>    RANGE   blue |
|---|---|---|
| The bicycle is blue. | The blue bicycle is expensive. | The fact that the bicycle was blue amused me. |

The shape in the left-hand column is an OBJECT described by an ATTRIBUTE. Given that shape in isolation—it is the entire idea to be expressed—the agent needs to create a copular sentence, which is a sentence with the verb *to be*. By contrast, when an OBJECT modified by an ATTRIBUTE is used as a case-role filler in another frame, as in the middle column, then a prenominal adjectival modifier is the default choice. Finally, when an ATTRIBUTE heads its frame and is used as a case-role filler in another frame, then a formulation like "the fact that N is Adj" is needed.

What we notice here is that the agent must first *detect* the shape of the overall meaning to be expressed and then figure out how to select and manipulate (using transformations) associated lexicon entries to create a sentence. This processing is carried out by two different kinds of code that wrap the shape of meaning, as illustrated in Figure 5.

The meaning representation to the upper left reflects what the agent wants to say. This example assumes that HUMAN-#20 is the LEIA's memory anchor for a man named Sam who is a boss, and that HUMAN-#25 is the memory anchor for a man named Harry. The needed shape of meaning is shown in the orange box. It covers the case when one human or intelligent agent (bot) wants, needs, requires, etc., another one to do something. Wanting, needing, and requiring are examples of MODALITY: e.g., the meaning *want* is MODALITY (TYPE volitive) (VALUE 1). Modalities scope over EVENTs that, themselves, can be of any shape. In our example, the EVENT takes an AGENT, a THEME, and an optional BENEFICIARY. The key feature of this shape is that the MODALITY is ATTRIBUTED-TO someone different than the AGENT of the EVENT (i.e., they are not coreferential). So, this shape can be used to generate the sentences shown in Fig. 2 (as well as many more), but it cannot be used to generate *John wants to <has to, must, is trying to* etc.*> fix the engine* since, in these, the modality is attributed to the same individual who is carrying out the event. The code called "Wrapper fit?" determines whether a given shape is applicable for processing the given meaning representation, and the code called "Apply wrapper" guides the agent in using the lexicon to express the meaning, which centrally includes modifying the basic, stored information using transformations.

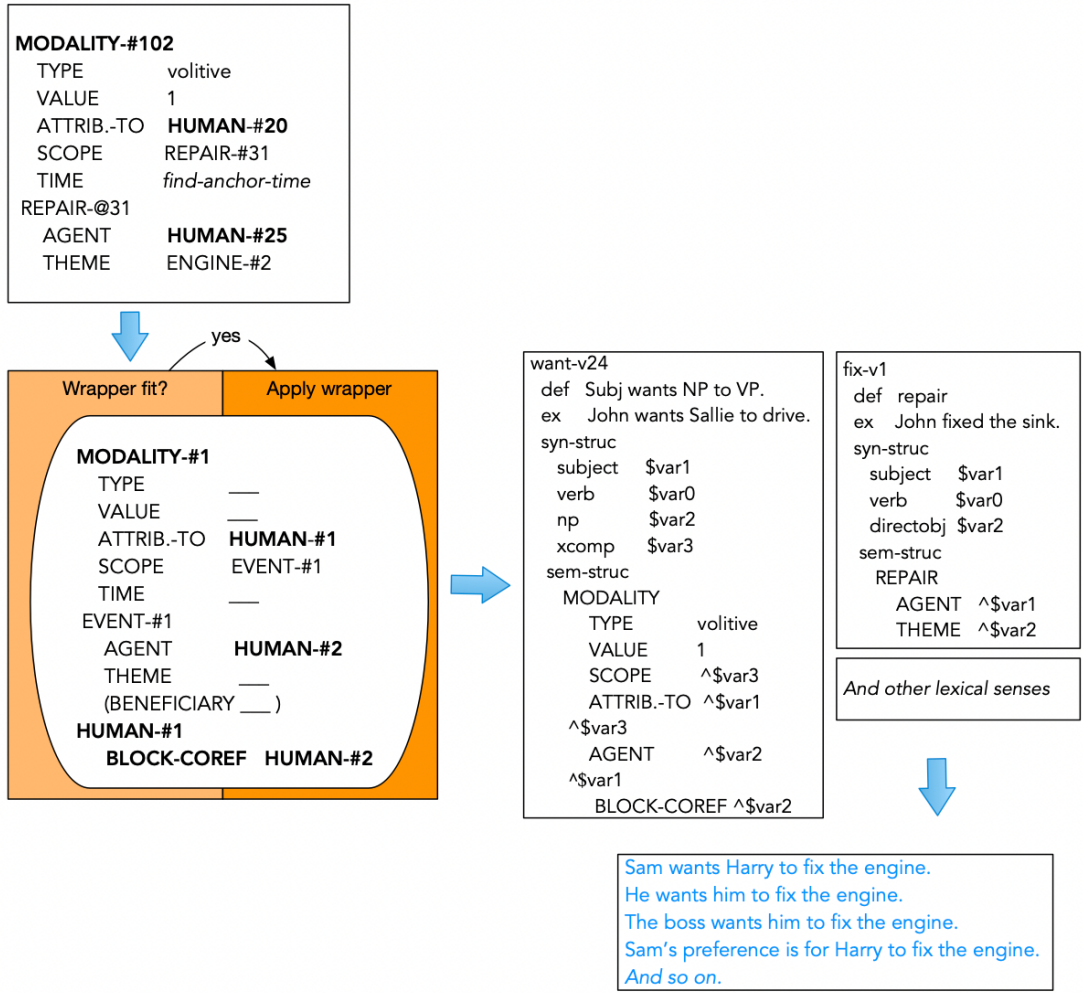

**Fig. 5.** Visualizing a shape of meaning (in the orange box) being used for language generation.

Further developing *shapes of meaning* involves positing shapes (deciding which frames to include, which are variable and fixed elements, etc.) and testing their combinability and generativity using ever more complex meanings. This is similar to the iterative testing, knowledge acquisition, and evaluation of an agent's language understanding capabilities.

Readers familiar with machine learning might wonder why we don't just have the agent learn the correspondences between meaning representations and sentences of English using a large corpus of such pairings—by analogy with machine translation. There are both practical and scientific reasons not to do so. The practical reason is that no such corpus exists and it would require a very large, expensive knowledge acquisition effort to acquire a large enough one. The scientific reason is that, even if the former could be done, it would not contribute to our understanding of human language processing, which is one objective of this program of work.

## 7. Shapes of Script Learning

Scripts are complex events, also known as procedural knowledge (Schank & Abelson, 1977). Formally, they are ontological concepts whose HAS-EVENT-AS-PART slot is filled by subevents with appropriately coreferenced participants and props. An example of a script is how to fill a gas tank, whose top-level frame is shown below. Each subevent (filler of HAS-EVENT-AS-PART) has its own property-rich frame, not shown here for reasons of space.

```
FILL-GAS-TANK
    IS-A                MACHINE-MAINTENANCE
    AGENT               HUMAN-OR-AGENT-#1
    THEME               GAS-TANK-#1
    CAUSED-BY           FLUID-LEVEL-#1          ; fuel level is low
    EFFECT              FLUID-LEVEL-#2          ; fuel tank is full
    HAS-EVENT-AS-PART   REMOVE-#1, INSERT-#1, PUMP-LIQUID-#1, REMOVE-#2, MOVE-#2, CLOSE-CONTAINER-#1
```

The idea is that you fill a gas tank because the gas level is low, and the result is that the tank is full. There are six ordered, top-level subevents: open the gas tank, remove the nozzle from the fuel dispenser, insert the nozzle into the gas tank, pump the gas until the tank is full, pull the nozzle out of the gas tank, return it to the fuel dispenser, and close the gas tank. In a robotic system, all of these events and their own subevents must be grounded in physical procedures that the agent's robotic systems can generate and its vision system can recognize.

Agent systems need to be able to learn scripts on the fly, whether the descriptions are provided by people through dialog, from language models, or from procedural documentation. However, script learning is not a single capability. On the one hand, scripts can range from simple to complex; and on the other hand, script learning relies on a large number of enabling capabilities. So, saying that an agent can learn scripts is too vague to be informative. The question is, what *exactly* can it do? Answering this requires decomposing script learning into its many component tasks and, for each one, identifying which actual phenomena the agent can handle at any given time. Both the top-level algorithm (Fig. 6) and the algorithms within each module involve shapes.

The gradient coloring in Fig. 6 is intended to convey that, in a given learning scenario, individual modules can present different challenge levels, with white being simple and dark blue being difficult. Fig. 7 shows how the different challenge levels of different learning scenarios can be visualized as different shapes of learning. In the left-hand scenario, understanding the input and figuring out what needs to be learned are difficult, but the agent is not expected to detect or fill lacunae in the information presented to it, and it is not required to describe back what it plans to learn (these modules are shown in gray: not applicable). By contrast, in the right-hand scenario, everything is easy except that the agent *is* supposed to detect and fill lacuna as well as describe back what it learned—both of which, for whatever reason, are difficult in the given scenario.

So, when agents are being evaluated for their ability to learn scripts, they are actually being evaluated for different things based on which challenges the scenario presents. Moreover, when teaching agents to be better learners, the best strategy is to focus on a subset of challenges at a time.

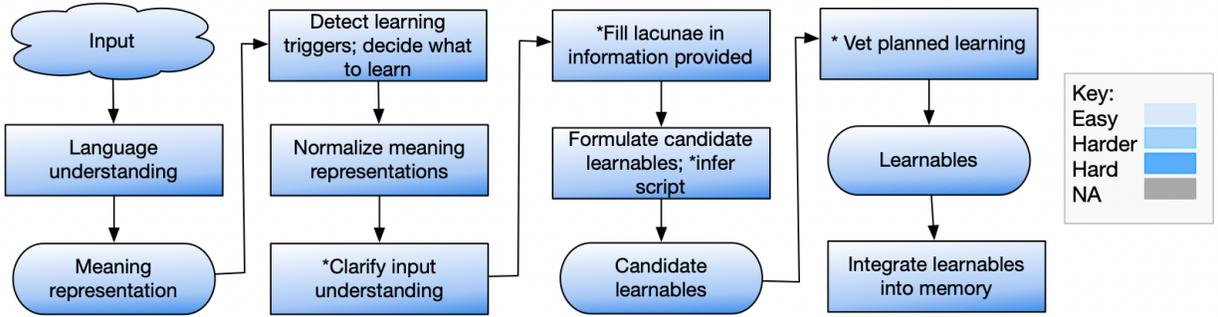

**Fig. 6.** Top level of the *shapes of learning* microtheory. Asterisks indicate optional modules.

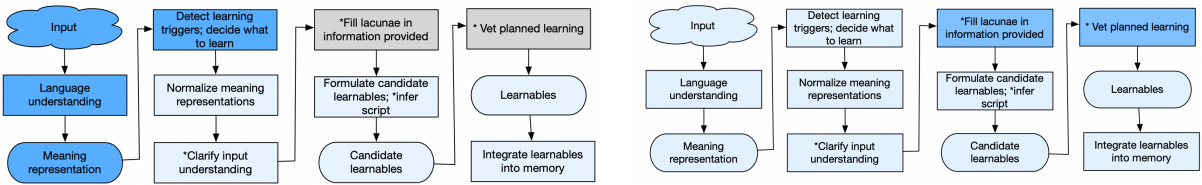

**Fig. 7.** Color coding helps to visualize different foci and challenges in script-learning scenarios. No need to read the box labels, just note the different shapes of learning reflected by the colors.

Shapes play out in the individual modules as well. We have already described this for language understanding and generation, which figure into multiple modules of the learning process. We will briefly mention select other examples.

1. Detecting what needs to be learned can range from easy to very difficult depending on whether the material is presented in a canonical shape.
    a. Easy: Here's how you [do something]. First [do A]. Then [do B]. And finally [do C].
    b. A little harder: [Doing X] requires [doing Y]. But first you have to make sure that [not Z].
    c. Hard: An cognitive robot is instructed to shadow a person all day and learn things. This involves understanding all of the events, which will be recorded in memory as *instances* of ontological concepts, and then determining if any sequences of them are prototypical and, therefore, should be stored as scripts in the ontology.
2. Clarifying input understanding through dialog (with a person or a language model) ranges in difficulty based on what, exactly, remains unclear and whether or not the agent needs to fully understand the input before acting.
    a. Easy: A single word or referring expression remains unclear, and the agent might even have a best guess based on the context.
    b. Harder: Multiple words, referring expressions, or speech acts remains unclear but the agent knows it needs to achieve full understanding, so it can work through the issues one by one.
    c. Hard: The agent has multiple problems understanding the input but it is authorized to act as long as it has reached an acceptable threshold of understanding. Determining what is actionable, and deciding what to do with the incompletely understood parts, can be quite challenging. (Cf. McShane et al., submitted, about assessing actionability.)
3. Detecting and filling lacunae in the information provided can also range from easy to hard.
    a. Easy: Agents know that they cannot learn a new concept without knowing its parent (its anchor it in the ontology), so if this information is lacking, it needs to be sought out.

b. Harder: If events in the script are presented using language that doesn't make their ordering clear (for example, "You have to X and Y" can imply *in that order* or in either order), this is a trigger for the agent to doublecheck.
   c. Hard: When people describe a complex event, they can imagine many things to be self-evident, such as that coffee beans need to be ground, and that you have to close the windows of a vehicle before washing it. Detecting what is left out can be difficult, and is a place where language models might help.

These examples should suffice to give an idea of how learning functionalities are graded, and how that grading is conceptualized by being closer or farther away from known, recorded shapes. A key takeaway is that, when an agent is learning to learn, and when we are evaluating its progress, we cannot expect it to deal with the most difficult phenomena in every module at the same time.

## 8. Final Thoughts

For reasons of space, this paper has presented only select examples of shapes in LEIA modeling, and ever so briefly. But there are other examples that are no less illustrative, such as:

1. **Shapes of coreference.** One of several strategies that LEIAs use to resolve coreference is recognizing predictive constellations of syntactic, semantic, and discourse features (LingAI, ch. 5; LongGame, ch. 5). For example, when direct objects in coordinated verb phrases have matching feature values (number, person, gender), they are highly likely to be coreferential, particularly if the first one is also a pronoun: *Patty grabbed <u>the cupcake</u> and scarfed <u>it</u> down; Patty grabbed <u>it</u> and scarfed <u>it</u> down*. Although this strategy does not account for all instances of referring expressions, it offers high-confidence resolutions for many, and it has proven useful cross-linguistically for even the challenging phenomena of ellipsis, 3$^{rd}$ person pronouns, and broad referring expressions.
2. **Shapes of dialog.** Dialog is modeled using adjacency pairs of ontological concepts: e.g., REQUEST-ACTION & RESPOND-TO-REQUEST-ACTION; PROPOSE-PLAN & RESPOND-TO-PROPOSED-PLAN. When the agent recognizes the interlocutor's communicative act (using a function recorded in the associated concept), it instantiates the concept's adjacency pair and follows *its* algorithm to guide the next move.
3. **Shapes of clarification.** When an agent doesn't fully understand a perceptual input, one of its options is to ask for clarification. How it does this reflects what it does and does not understand in the situation. For example, if an agent understands everything except for the referent of *he,* it might ask "Phil or Hal?", whereas if there are many types of residual ambiguity, it might paraphrase what it thinks was meant and ask if its understanding is correct. This decision-making is an example of shapes-based modeling in that it anticipates typical cases and has the agent respond in what people would consider the most efficient and expected way.
4. **Shapes of actionability.** In some applications, it is acceptable and even necessary for agents to act despite incomplete situational understanding and/or confidence in their reasoning. Assessing actionability could be intractably complicated, but it is not when agents are guided by typical feature constellations and eventualities, using an algorithm detailed in McShane et al. (Submitted).
5. **Shapes of control.** The robotic (tactical) layer of HARMONIC (Oruganti et al. 2024a, 2024b) is conceptualized using shapes, where some actions are treated as non-decomposable monoliths and others are decomposable into structured control primitives defined by ontological scripts. We theorize *shapes of control* as cognitively inspired, parameterized control schemas that translate qualitative human instructions into quantitative robotic actions. Grounded in control theory, each shape encodes reference values, control parameters, and error signal quantifications, supporting both feedback and feedforward mechanisms. These modular units enable the system to dynamically compose, tune, and stabilize behaviors in response to human requests, intent, and environmental context, forming a critical interface between high-level reasoning and low-level robotic control and actuation in the HARMONIC architecture.

6. **Shapes under the hood.** LEIA systems include dynamically populated under-the-hood panels that show human-interpretable traces of system processing, presented using the kinds of knowledge structures described earlier (LingAI, section 8.1.5; LongGame, section 8.7; Nirenburg et al., 2024). These have been invaluable for demonstrating system capabilities to a wide variety of audiences.

7. **Diagrams as shapes**. In order to develop agent systems over time without losses, knowledge engineers and system engineers need to sign off on a grain-size of algorithm that stretches both beyond their comfort zones. This is because implementation decisions that might initially seem unimportant can have serious consequences for an algorithm's extensibility. We use graphic toolsets to create visualizations of these algorithms, as illustrated by the algorithm for language understanding available at https://homepages.hass.rpi.edu/mcsham2/Appendix-Long-Game/APP-Long-Game-NLU.pdf. In visualizing algorithms we are guided, though not limited, by best practices of UML documentation (LongGame, section 2.5.2).

To reiterate a key point from earlier: *shapes* is not simply a new word to refer to a hodgepodge of related concepts that have been discussed for a half century. Instead, it is a specific approach to cognitive modeling that allows all new models to be variations on a well-established theme. The shapes orientation says: Let's focus on typical cases first, since that will go a long way to making agents useful; Let's make recovery strategies as generic as possible, particularly exploiting the fact the humans are in the loop; Let's think about models as pictures, be they diagrams or templates, so that we can clearly understand, explain and remember them; Let's evaluate systems based on what, exactly, we expect them to cover at any given time; and Let's remember that *shapes* are not only about engineering, they capture principles of human cognition that provide insights into how it is possible to function effectively in a complex, open-ended world.

**Acknowledgements**

This research was supported in part by Grants N00014-16-1-2118 and N00014-17-1-2218 from the U.S. Office of Naval Research. Any opinions or findings expressed in this material are those of the authors and do not necessarily reflect the views of the Office of Naval Research.**References**

Bybee, J. L. (2013). Usage-based theory and exemplar representations of constructions. In T. Hoffmann & G. Trousdale (Eds.), *The Oxford Handbook of Construction Grammar* (pp. 49–69). Oxford University Press.

Chomsky, N. (1957). *Syntactic Structures*. Mouton.

Connell, L., & Lynott, D. (2024). What can language models tell us about Human Cognition? *Current Directions in Psychological Science*, *33*(3), 181-189.

Cuskley, C., Woods, R., & Flaherty, M. (2024). The limitations of large language models for understanding human language and cognition. *Open Mind* 8: 1058-1083.

Fillmore, C. J., & Baker, C. F. (2009). A frames approach to semantic analysis. In B. Heine & H. Narrog (Eds.), *The Oxford Handbook of Linguistic Analysis* (pp. 313–340). Oxford University Press.

Gentner, D. & Smith, L. A. (2013). Analogical learning and reasoning. In D. Reisberg (Ed.), *The Oxford Handbook of Cognitive Psychology*, 668–681. New York, NY: Oxford University Press.

Hoffmann, T., & Trousdale, G. (Eds.). (2013). *The Oxford Handbook of Construction Grammar*. Oxford University Press.15